\documentclass{article}
\usepackage{spconf,amsmath,graphicx}
\usepackage{booktabs}
\usepackage{amssymb}
\usepackage{etoolbox}
\usepackage{setspace}
\usepackage[colorlinks=true, linkcolor=black, citecolor=black, urlcolor=blue]{hyperref}

\title{Deep Active Learning with Manifold-Preserving Trajectory Sampling}
\name{\begin{tabular}{c}Yingrui Ji$^1$,
        Vijaya Sindhoori Kaza$^2$,
        Nishanth Artham$^3$, 
        Tianyang Wang$^1$\sthanks{corresponding author}, 
        \end{tabular}
        }
\address{$^1$The University of Alabama at Birmingham ~~~~~~~
$^2$HPE  ~~~~~~~
$^3$Chaitanya Bharathi Institute of Technology
}
\vspace{-3em}

\begin{document}
%
\maketitle

\begin{abstract}
\vspace{-0.2em}
Active learning (AL) is for optimizing the selection of unlabeled data for annotation (labeling), aiming to enhance model performance while minimizing labeling effort. The key question in AL is which unlabeled data should be selected for annotation. Existing deep AL methods arguably suffer from bias incurred by clabeled data, which takes a much lower percentage than unlabeled data in AL context. We observe that such an issue is severe in different types of data, such as vision and non-vision data. To address this issue, we propose a novel method, namely Manifold-Preserving Trajectory Sampling (MPTS), aiming to enforce the feature space learned from labeled data to represent a more accurate manifold. By doing so, we expect to effectively correct the bias incurred by labeled data, which can cause a biased selection of unlabeled data. Despite its focus on manifold, the proposed method can be conveniently implemented by performing distribution mapping with MMD (Maximum Mean Discrepancies). Extensive experiments on various vision and non-vision benchmark datasets demonstrate the superiority of our method. Our source code can be found \underline{\href{https://drive.google.com/file/d/1pbY2OGRbAIVIyfY7MTiPJpy5xRQrx2X3/view?usp=sharing}{here}}. 

\end{abstract}
\vspace{-0.8em}
\section{Introduction}
\vspace{-0.5em}
\label{sec:intro}

Active learning (AL) has emerged as an effective strategy to enhance model performance while minimizing the need for extensive labeled data. The fundamental concept of active learning \cite{cohn1996active} lies in allowing the model to selectively choose the data it learns from, thereby achieving significant performance improvements with fewer labeled samples. 


Typical uncertainty-based AL solutions include techniques such as Monte Carlo (MC) dropout \cite{gal2016dropout}, which estimates uncertainty by performing multiple stochastic forward passes with dropout layers enabled. The assumption behind this is to improve uncertainty estimation by sampling parameters from their posterior distribution. Another approach Bayesian Neural Network (BNN) \cite{gal2017deep} models the posterior distribution over the parameters directly, often assuming a Gaussian form. Other solutions involve model ensemble \cite{beluch2018power,czarnecki2015adaptive}, which trains multiple independent instances of a neural network with different initialization. While these methods provide improved estimates of uncertainty, they often suffer from two fundamental limitations as follows. 

First, to accurately estimate the posterior distribution of model parameters $P(\theta|D)$, it is crucial to have a representative distribution of the data. However, current active learning methods often face the risk of biased data sampling due to the limited number of examples used for model training. This issue becomes more pronounced in a multi-cycle active annotation process, where the repeated selection of the most uncertain examples can lead to a progressively biased data distribution. Second, many existing approaches rely on explicit assumptions, such as assuming a Gaussian form for the posterior distribution, or require modifications to the model architecture, such as incorporating dropout layers, to enable parameter sampling. These assumptions and architectural changes may not always be reliable, potentially limiting the applicability of the methods to different models. 


To address these challenges, we introduce a novel active learning algorithm named Manifold-Preserving Trajectory Sampling (MPTS). A core idea in our approach is to ensure the model is trained to be consistent with the true data manifold, thereby mitigating the risk of bias that arises from only using the most uncertain examples across multiple active learning cycles. When training the discriminative model for uncertainty estimation, we utilize the sufficient unlabeled examples to regularize the feature distribution from the labeled examples. 
During the regularized training, we sample parameters from the optimization trajectory near local minima by periodically averaging model parameters encountered during the latter stages of training. This technique provides an effective sampling by considering multiple points along the trajectory that are close to a local minimum, thereby capturing a diverse set of model parameters that represent different modes of the posterior distribution.

To our best knowledge, our paper is the first to propose data bias correction in estimating the posterior distribution $P(\theta|D)$ for active learning. Besides, while parameter sampling from optimization trajectories has been explored in the literature, these methods often rely on explicit distribution modeling, which can increase complexity and introduce the risk of making unreliable assumptions. Our approach ensures an effective and unbiased parameter sampling from the optimization trajectory fir uncertainty estimation. 
Through extensive experiments, we demonstrate that our method consistently outperforms various state-of-the-art active learning methods on multiple datasets covering images and tabular data. 
This highlights the broad applicability and reliability of our method.


\vspace{-1em}
\section{Related Work}
\vspace{-0.8em}
\subsection{Active Learning}
\vspace{-0.5em}
Active learning is a pivot research area in machine learning, focused on optimizing data annotations to enhance model performance with fewer labelled samples. Most AL methods mainly consider uncertainty as a crucial criterion to intelligently sample data that improves model's generalization. Such methods prioritize data points with high prediction variance or near the decision boundary, employing techniques like MC-Dropout \cite{gal2016dropout}, Query-by-Committee (QBC) \cite{gorriz2017}, and adversarial training \cite{ducoffe2018} to address overconfident deep neural networks \cite{Tong2001, Sinha_2019_ICCV}. Influence-based AL approaches select data points based on their estimated impact on model performance, using schemes like Learning Loss \cite{Yoo_2019_CVPR}, and the Influence Function \cite{koh17a} that leverages gradient to estimate changes in prediction accuracy \cite{Liu_2021_ICCV, Wang_2022}. Besides, BADGE \cite{Ash_2021} also aims to select uncertain data by evaluating gradient. Many deep AL methods resort to auxiliary models to estimate data uncertainty. Typical works include VAAL \cite{Sinha_2019_ICCV} that uses an auxiliary auto-encoder, and GCNAL \cite{Caramalau_2021_CVPR} that employs a graph network as the auxiliary model. Unlike these methods, the Coreset \cite{sener2018} is free of any auxiliary models, but suffering from a slow optimization process (e.g., solving a classical K-center or 0-1 Knapsack problem) during data selection. Several other works, such as \cite{zhang2020state}, rely on complicated training fashion (e.g., adversarial), and it will be challenging if using such methods on a different data format (e.g., 3D medical images of voxels) other than 2D natural images.



\vspace{-1.3em}
\subsection{Posterior Approximation for Bayesian Neural Networks}
\vspace{-0.5em}
Bayesian Neural Networks (BNNs) are designed to provide robust uncertainty estimates by treating the network’s parameters as probabilistic distributions rather than fixed values. This approach is essential for capturing uncertainty in tasks like active learning. Several works \cite{Maddox2018, Maddox_2019, Linden2020} propose to estimate posterior distributions by averaging the training checkpoints. To this end, they use Stochastic Weight Averaging (SWA) to perform the averaging operation, improving the uncertainty estimation. These methods offer practical solutions for reliable uncertainty estimation in deep networks. Notably, our method has a very low level of similarity with these methods, as we propose a brand new solution to estimate posterior considering both labeled and unlabeled data simultaneously.

\vspace{-1em}
\section{Method}
\vspace{-0.8em}
\subsection{Preliminaries}
\vspace{-0.5em}
We first introduce the multi-cycle deep active learning problem setting. Starting with a set of unlabeled samples, $\mathcal{U}$, and a labeled set, $\mathcal{L}$, the objective is to select a subset from $\mathcal{U}$ according to a predefined annotation budget. This chosen subset, ${X_N}$, is annotated by experts or equivalent sources, resulting in labeled data: $\mathcal{L} \leftarrow \mathcal{L} \cup \{X_N, Y_N\}$, where ${Y_N}$ are the labels. The remaining unlabeled data is updated: $\mathcal{U} \leftarrow \mathcal{U} \setminus {X_N}$. 
In the process of subset selection, we typically need to train a model parameterized with ${\theta}$ as $f(\cdot;\theta)$. Considering a classification task with $C$ classes, Entropy can be used to estimate the prediction uncertainty for a given example $x$ as 
\begin{equation}
    H(x) = -\underset{c \in C}{\Sigma}p(y=c|x,\theta) \log p(y=c|x,\theta).
\label{eq:entropy}
\end{equation}
where $p(y=c|x,\theta)$ corresponds to the softmax probability of the $c$-th class from the prediction $f(x;\theta)$. 

For better estimation of the probability, BNN approaches incorporate the posterior distribution of $\theta$ by
\begin{equation}
    p(y=c|x,D) = \int p(y=c|x,\theta) p(\theta|D) d\theta.
\end{equation}
where $D$ represents the data distribution. While $p(y=c|x,\theta)$ can be easily calculated from the network, the key problems turn into (1) estimating the posterior distribution $p(\theta|D)$ and (2) sampling from the parameter distribution to approximate the integration. Next, we introduce our solutions for them. 

\vspace{-0.8em}
\subsection{Posterior Estimation with Manifold-Preserving} 
Traditional MC-dropout \cite{gal2016dropout} or other BNN based approaches suggest to estimate the posterior distribution $p(\theta|D)$ by training a model with the available labeled set $\mathcal{L}$. Here we analyze its potential risk and propose our method by the following analysis. 

Since a direct calculation of $p(\theta|D)$ is infeasible, we apply the Bayes rule as
\begin{equation}
    p(\theta|D) = \frac{p(\theta) p(D|\theta)}{p(D)} \propto p(\theta) p(D|\theta).
\label{eq:posterior}
\end{equation}
Regarding $p(D)$ as data sampling, we further decompose the posterior distribution into the multiplication of the prior term $p(\theta)$ and the likelihood $p(D|\theta)$. Since $p(\theta)$ can be realized by common regularization techniques such as weight decay, the flexible task is to estimate $p(D|\theta)$. Given the data distribution $D$, we convert the problem of sampling the most possible parameters into finding those $\theta$ that maximize the likelihood term.

Denoting the input data by $X$, labels by $Y$ and the deep features by $Z$ with the dependency chain as $X \rightarrow Z \rightarrow Y$ in a discriminative model, we further decompose the likelihood as $p(D|\theta) = p(X,Z,Y|\theta)$. Therefore we have
\begin{equation}
    p(D|\theta) = p(X|\theta)p(Z|X,\theta)P(Y|X,Z,\theta).
\end{equation}
Given the potentially biased labeled set $\mathcal{L}$ as the training set, existing approaches only have the effect of maximizing the conditional distribution term $p(Y|X,Z,\theta)$, which is sub-optimal in maximizing $p(D|\theta)$. 

While label information is only available for $\mathcal{L}$, we propose to correct the data bias by regularizing the second term $p(Z|X,\theta)$ with real distribution calculated by $\mathcal{L} \cup \mathcal{U}$. In this way, we enforce the feature space $Z$ learned from biased input $X$ to represent a more accurate manifold. Specifically, denoting $f_e$ as the feature extractor part of the neural network $f$, we aim to align the distribution $Z_{\mathcal{L}}=f_e(X_{\mathcal{L}}; \theta)$ with $Z_* \approx f_e(X_{\mathcal{L} \cup \mathcal{U}}; \theta)$. 

To achieve this, we adopt Maximum Mean Discrepancies $\mathrm{MMD}$~\cite{borgwardt2006integrating}, which is originally designed to check whether two samples are from the same distribution. Denoting $\mathcal{H}$ as a class of functions $h : \mathcal{X} \to R$, $\mathrm{MMD}$ is defined as
\begin{equation}
    \mathrm{MMD}(Z_{\mathcal{L}}, Z_*) = \sup_{h \in \mathcal{H}}\{\mathop{\mathbb{E}}_{z\sim Z_{\mathcal{L}}}[h(z)] - \mathop{\mathbb{E}}_{z\sim Z_*}[h(z)]\}.
\label{eq:mmd}
\end{equation}
By specifying $\mathcal{H}$ as a reproducing kernel Hilbert space, Eq. (\ref{eq:mmd}) can be estimated by comparing the square distance between the empirical kernel mean embeddings as
\begin{equation} 
\begin{aligned}
\mathrm{MMD}^2(Z_{\mathcal{L}}, Z_*) \approx \| \frac{1}{|Z_{\mathcal{L}}|}\sum_{z \in Z_{\mathcal{L}}} \kappa(z) - \frac{1}{|Z_{\mathcal{L} \cup \mathcal{U}}|}\sum_{z \in Z_{\mathcal{L} \cup \mathcal{U}}} \kappa(z) \|^2. 
  \end{aligned}
\end{equation}
Here we follow existing transfer learning literature by expanding the quadratic term and calculate the kernel functions corresponding to $\kappa$, for which a Gaussian radial basis function (RBF) is adopted. In practice, two  batch of examples randomly sampled from $\mathcal{L}$ and $\mathcal{L} \cup \mathcal{U}$ are used to minimize the MMD distance in each training iteration. 

To obtain optimal parameters, we simultaneously minimize the standard supervised error $L_{ce}$ (e.g., cross-entropy loss) and the manifold preserving loss as
\begin{equation}
    L = L_{ce}(X_{\mathcal{L}}) + \lambda \mathrm{MMD}^2(Z_{\mathcal{L}}, Z_*).
\label{eq:loss}
\end{equation}

\vspace{-1em}
\subsection{Sampling from Optimization Trajectory}
We use the optimization trajectory to sample a diverse set of checkpoints by leveraging the path of minimizing Eq. (\ref{eq:loss}) with stochastic gradient descent (SGD). This approach intends to explore different regions of the parameter space that are likely to have a high posterior probability as in Eq. (\ref{eq:posterior}). Specifically, we adopt the technique proposed by Stochastic Weight Averaging (SWA) for parameter sampling. By using a cyclic learning rate after the half convergence, a set of checkpoints $\{\theta_{t_i}\}_{i=1}^n$ are sampled and saved. 

Finally, we calculate the prediction probability by averaging the prediction of each sampled parameter as
\begin{equation}
    p(y=c|x,D)= \frac{1}{n} \Sigma_{i=1}^n p(y=c|x,\theta_{t_i}), 
\label{eq:swa}
\end{equation}
where each $\theta_{t_i}$ is sampled in the trajectory of minimizing Eq. (\ref{eq:loss}). We substitute Eq. (\ref{eq:swa}) into Eq. (\ref{eq:entropy}) to calculate the uncertainty for each unlabeled example from $\mathcal{U}$. 

Note that the manifold preserving and parameter averaging approaches are designed only to select uncertain examples in our active learning framework, despite that SWA is more popular as a strategy to improve model accuracy. 

\vspace{-0.7em}
\section{Experiments}
\vspace{-0.5em}
Here we introduce a series of experiments conducted to validate the proposed method. To make the evaluation more comprehensive, we consider multiple datasets, backbone models, as well as different AL settings. We refer readers to Table \ref{tab:headings} for details.  
\vspace{-1em}
\subsection{Datasets and Baselines}
\textbf{Datasets.}~As most deep AL methods have been evaluated in computer vision challenges, we also adopt four widely used benchmark vision datasets to evaluate our method, including MNIST \cite{24MNIST}, CIFAR10 \cite{22CIFAR}, SVHN \cite{30netzer2011reading}, and Mini-ImageNet \cite{34MiniImageNet}. In addition, we also incorporate two typical non-vision datasets for the evaluation, namely OpenML-6 \cite{41openml} and OpenML-155 \cite{41openml}, which are tabular datasets from the OpenML repository, including structured data with mixed types of features. 

\noindent \textbf{Baselines.}~The baseline methods that we consider in this paper can be categorized into three groups. The first group resorts to estimate data uncertainty based on posterior, including Entropy \cite{40Entropy}, BALD \cite{gal2017deep}, BADGE \cite{3MLP}. The second group designs customized methods to evaluate data uncertainty, including Coreset \cite{35sener2018}, CDAL \cite{2agarwal2020contextual}, and Feature Mixing \cite{Parvaneh_2022_CVPR}. The third group relies on auxiliary models and/or special training fashion (e.g., adversarial), including Adversarial Deep Fool \cite{ducoffe2018} and GCNAL \cite{Caramalau_2021_CVPR}. In addition, Random selection is also included as it is a straightforward yet effective method in several scenarios. 

\noindent
\textbf{Models.}~We use three types of deep models as the backbones. Specifically, we use MLP \cite{3MLP} for MNIST, ResNet-18 \cite{17ResNet18} as a typical CNN for CIFAR10 and SVHN, and  vision transformer (ViT) \cite{12visualtransformer} as a typical foundation model for Mini-ImageNet. We also use MLP for the two non-vision datasets.


\subsection{Experimental Settings}
For each dataset, following the common practice in AL literature, we randomly select a small portion of data as initial samples and annotate them. The number of such such samples is 100 for all the datasets, except the Mini-ImageNet in which we use 1000 initial samples. Then in each AL round (covering both model training and data selection phases), we select 100 unlabeled samples (labeling budget) for all the datasets, except the Mini-ImageNet where we select 1000 unlabeled samples for initial annotation. When MLP or CNN is used as the backbone, within each AL round, we train the model for 100 epochs. When pre-trained ViT is used, we fine-tune it for 1000 epochs within each AL round. We adopt a learning rate of ${1e-3}$ for vision datasets and ${1e-4}$ for non-vision datasets. The batchsize is set to 64 for all the experiments. Notably, we train the MLP and CNN from scratch, whereas we fine-tune the pre-trained ViT following the practice in \cite{Parvaneh_2022_CVPR} for a fair comparison. To reduce randomness, we repeat each experiment for 5 times and average the results as the final one.


\begin{table*}[tb]
  \caption{A summary of various AL settings we use in the experiments. 
  }
  \label{tab:headings}
  \centering
  \vspace{0.5em}
  \resizebox{.8\textwidth}{!}
  {
  \begin{tabular}{@{\hskip 0.1in}c@{\hskip 0.1in}c@{\hskip 0.1in}c@{\hskip 0.1in}c@{\hskip 0.1in}c@{\hskip 0.1in}c@{\hskip 0.1in}c@{\hskip 0.1in}c@{\hskip 0.1in}c@{}}
    \toprule
        Dataset & Pool Size & Label Size & Input & Initial Instances & Budget & Backbone & Initialization\\
    \midrule
        CIFAR10 \cite{22CIFAR}& 50,000 & 10 & 32 × 32 & 100 & 100 & ResNet-18& Random \\
        MNIST \cite{24MNIST} & 50,000   & 10 & 28 × 28 & 100 & 100 & MLP & Random\\
        SVHN \cite{30netzer2011reading} & 50,000   & 10 & 32 × 32 & 100 & 100 & ResNet-18 & Random \\
        Mini-ImageNet \cite{34MiniImageNet} & 48,000 & 100 & 84 × 84 & 1000 & 1000 & ViT-Small & Pre-trained\\
    \midrule
        OpenML-6 \cite{41openml} & 18,000 & 26 & 16 & 100 & 100 & MLP & Random\\
        OpenML-155 \cite{41openml} & 50,000 & 9 & 10 & 100 & 100 & MLP & Random\\
    \bottomrule
  \end{tabular}
  } 
\vspace{-1.5em}
\end{table*}


\begin{figure}[t]
    \centering
    \includegraphics[width=0.49\linewidth]{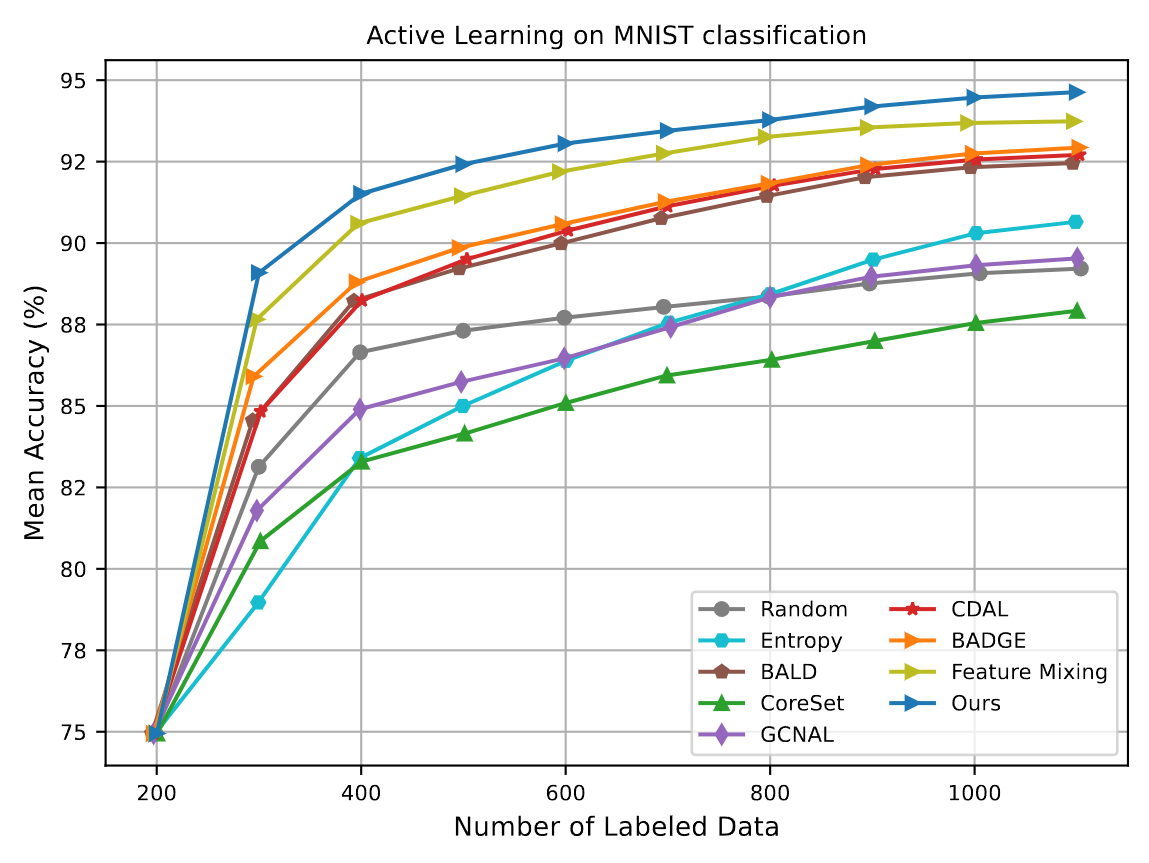}
    \includegraphics[width=0.49\linewidth]{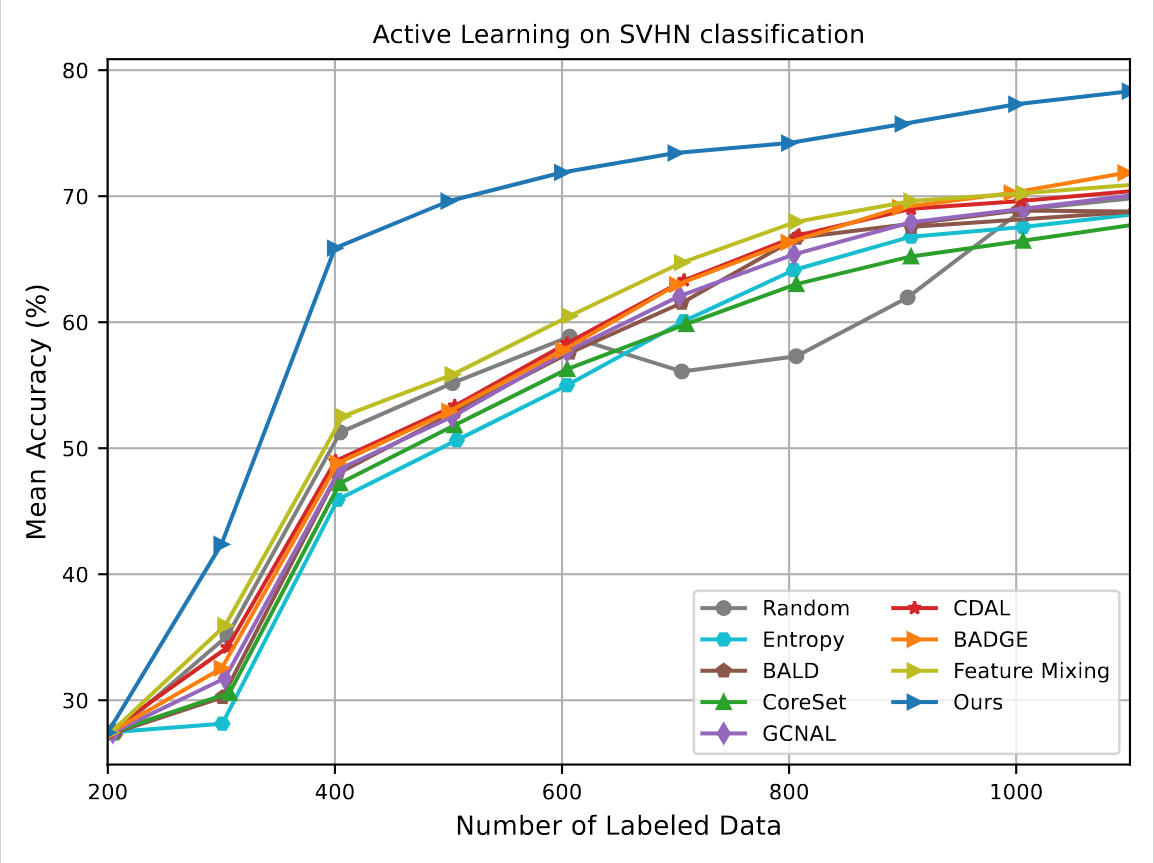}
    \includegraphics[width=0.49\linewidth]{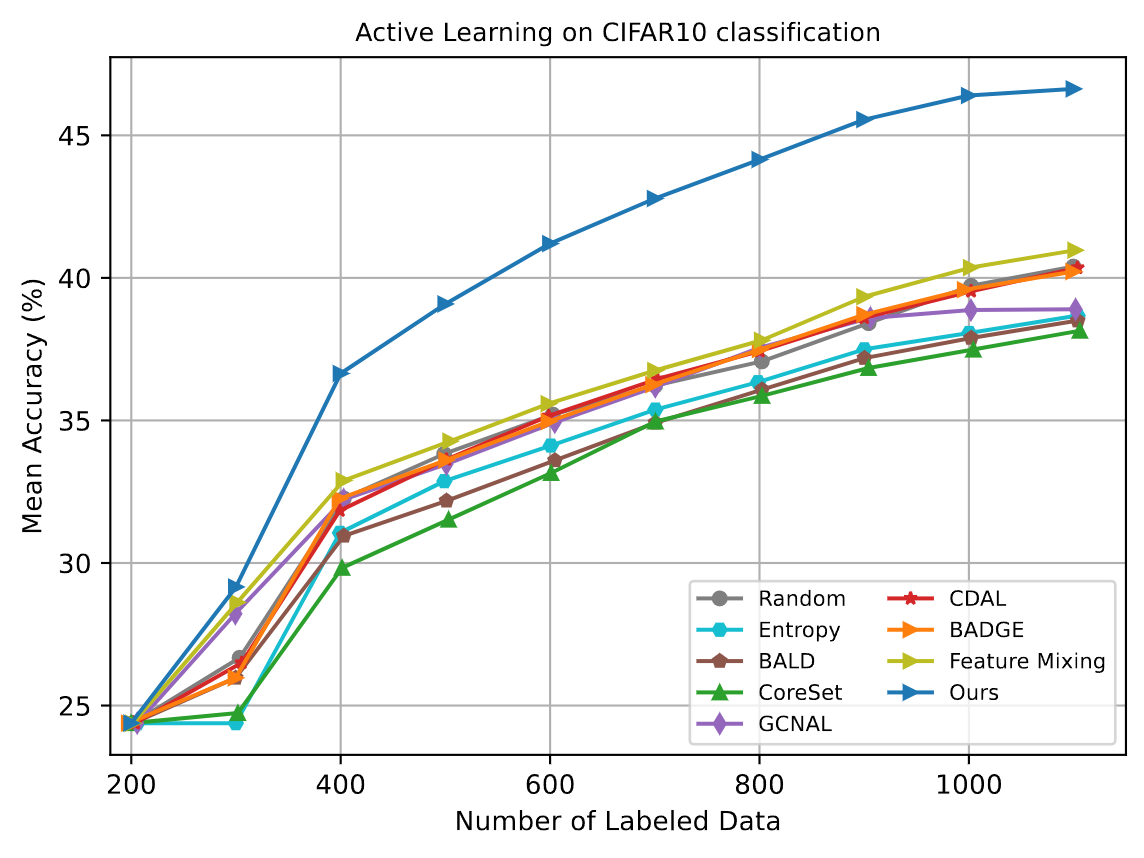}
    \includegraphics[width=0.49\linewidth]{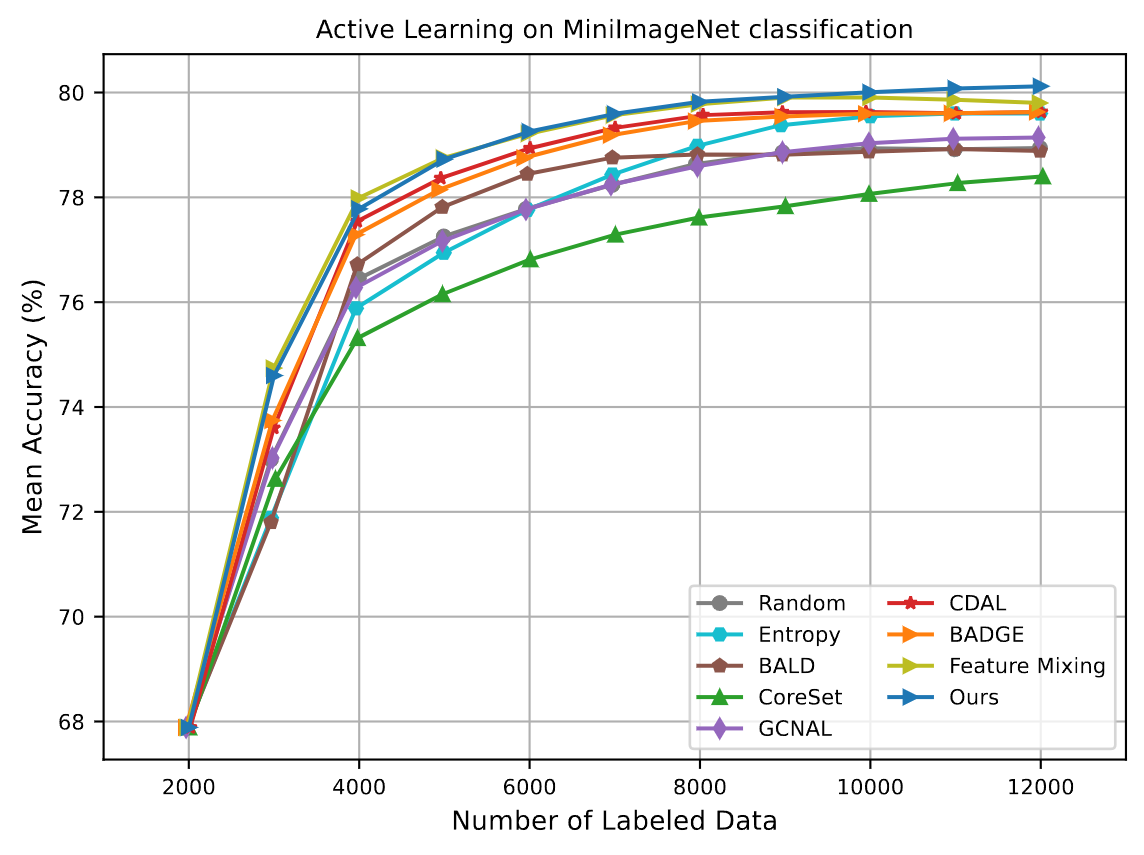}
    \caption{Performance comparison of the methods on vision datasets. Zoom in for a better view. 
    }
    \label{fig:res}
    \vspace{-0.3em}
\end{figure}

\subsection{Results and Analysis}
\vspace{-0.5em}

As shown in Figure \ref{fig:res}, across the multiple datasets widely used in computer vision scenarios, the proposed method outperforms the others in terms of the classification accuracy. In addition to this overall evaluation, we have the following observations. First, our method yields solid results when different backbone models are employed, i.e., MLP for MNIST, CNN for CIFAR10 and SVHN, and ViT for Mini-ImageNet. This observation demonstrates the strong adaptability to the mainstream deep learning models. Second, when labeling budget is limited, such that only 100 samples can be annotated at a time (i.e., MNIST, CIFAR10, and SVHN), our method gives consistent better performance than the others, demonstrating its potential in real-world scenarios where labeling cost could be extremely high (e.g., in medical domain). Third, compared to the improvement in MNIST, our method outperforms the others by a more significant margin in CIFAR10 and SVHN, indicating the capability of our method on handling data of complex scene (i.e., CIFAR10 and SVHN have more complex scene than MNIST). Fourth, when labeling budget is relatively high (i.e., 1000 in Mini-ImageNet whereas 100 in the others), our method still shows consistent superiority over the others. This further demonstrates the strong capability of our method since most AL peers fail to show their superiority when labeling budget is high. Last, compared to the AL method that is based on auxiliary models, i.e., GCNAL \cite{Caramalau_2021_CVPR} that uses an auxiliary graph network for data selection in addition to the task model, our method yields more appealing results, suggesting that developing deep AL methods without leveraging auxiliary models could be a promising effort. 

\begin{figure}[t]
    \centering
    \includegraphics[width=0.49\linewidth]{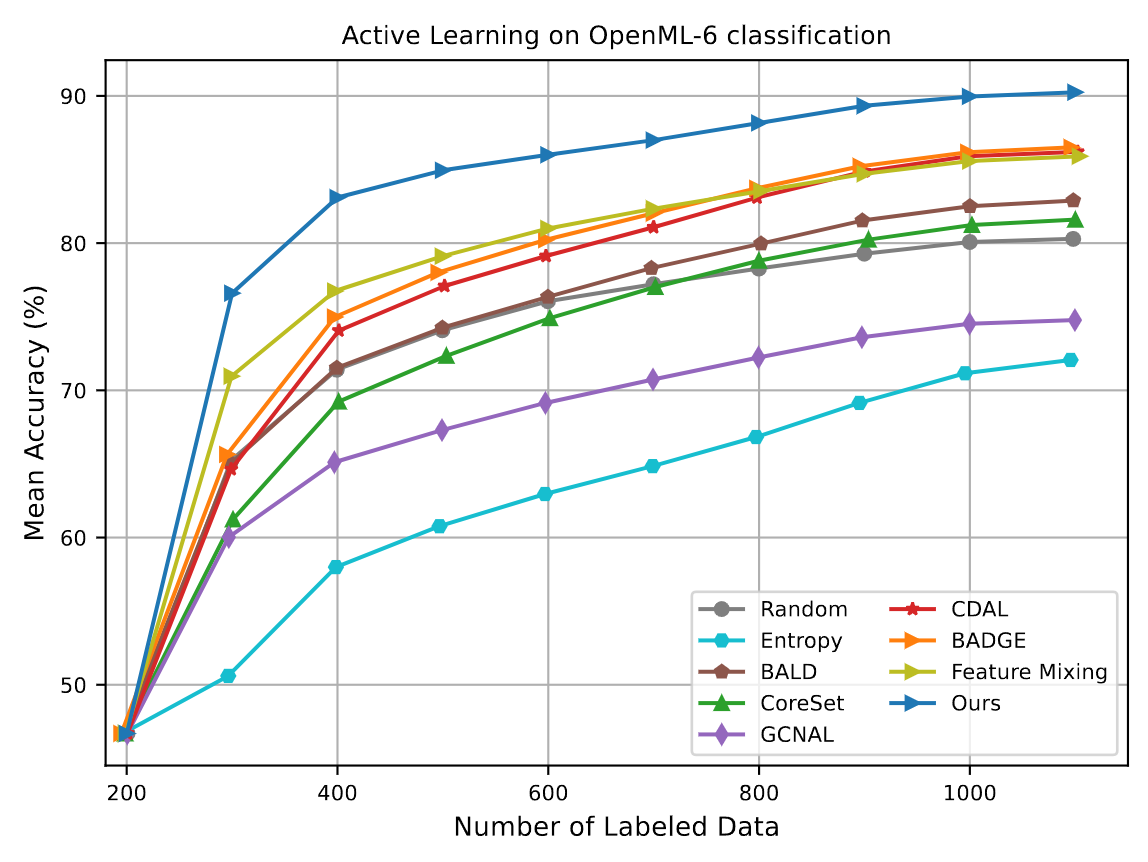}
    \includegraphics[width=0.49\linewidth]{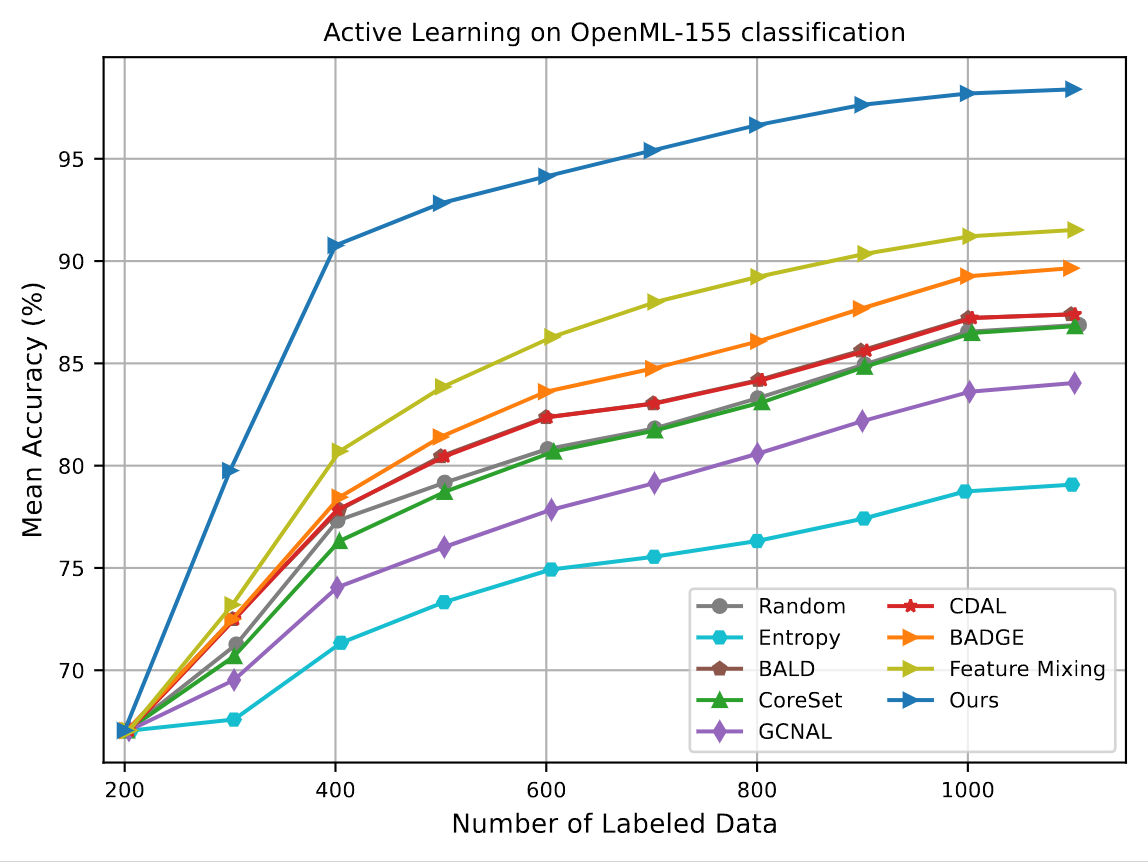}
    
    \caption{Performance comparison of the methods on non-vision datasets. Zoom in for a better view. 
    }
    \label{fig:res-nonvision}
\end{figure}

In addition, as illustrated in Figure \ref{fig:res-nonvision} we surprisingly find that our method significantly outperforms the others in non-vision datasets, such as OpenML-6 and OpenML-155. As MLP is used as the backbone model rather than CNN for such structured tabular data, the inferior performance yielded by the other methods indicate that MLP is more impacted by data bias, while our method can effectively handle the bias issue. Notably, in MNIST where MLP is also used (see Fig. \ref{fig:res}), the performance gaps between our method and the others are not that big since MNIST is a fairly simple dataset that cannot make the methods show their upper bounds.

\section{Conclusion}
In this paper, we theoretically find that it is risky to estimate data uncertainty based on estimating  posterior distribution only with labeled data in deep active learning. Motivated by this, we propose a novel substitution leveraging manifold-preserving to avoid the risk. We then design a simple and feasible solution to integrate the proposed scheme into the training of deep models within active learning context. Experimental results demonstrate that the proposed method is superior in various scenarios. Moreover, the simplicity of the realization may lead to its potential of being widely used in active learning tasks and beyond.


\clearpage
\makeatletter
\patchcmd{\thebibliography}{\labelsep}{\labelsep \setlength{\itemsep}{-3pt}}{}{}
\makeatother

\begingroup 
\bibliographystyle{IEEEbib}
\bibliography{strings,refs}
\endgroup
\end{document}